\documentclass[11pt]{article}

\usepackage[T1]{fontenc}
\usepackage[utf8]{inputenc}
\usepackage{lmodern}
\usepackage[letterpaper,margin=1in]{geometry}
\usepackage{amsmath,amssymb}
\usepackage{booktabs}
\usepackage{array}
\usepackage{graphicx}
\usepackage{xcolor}
\usepackage[font=small,labelfont=bf]{caption}
\usepackage[numbers,sort&compress]{natbib}
\usepackage{microtype}
\usepackage{enumitem}
\usepackage{float}
\usepackage[section]{placeins}
\usepackage[hidelinks]{hyperref}

\setlist{itemsep=2pt,topsep=4pt}
\newcommand{\figref}[1]{Figure~\ref{#1}}
\newcommand{\tabref}[1]{Table~\ref{#1}}
\newcommand{\secref}[1]{\S\ref{#1}}
\newcommand{\tablebox}[1]{\resizebox{\ifdim\width>\linewidth\linewidth\else\width\fi}{!}{#1}}

\title{\textbf{Similar Scores, Opposite Failures:}\\[2pt]
\large Human Correction as an Instrument for Video Caption Evaluation}
\author{Ankit Khedia\quad\ Astitva Jaiswal\quad\ Chaitanya Prasad\quad\ Anisha Raju\quad\ Aangeeras\\[4pt]
Deccan AI\\[2pt]
\small\texttt{ankit.khedia@deccan.ai}\quad\texttt{astitva.jaiswal@deccan.ai}\quad\texttt{ext.chaitanya@deccan.ai}\\
\small\texttt{ext.anisha@deccan.ai}\quad\texttt{aangeeras@deccan.ai}}
\date{October 2026}

\begin{document}
\maketitle

\begin{abstract}
Video captioning models are increasingly evaluated with metrics designed for single-sentence image captions. We test whether those metrics recover human judgement on dense, multi-window video description, and find that they do not. Six frontier models captioned 196 short clips. Trained annotators post-edited every caption against a failure taxonomy fixed before annotation, once as a whole-clip overview and once window by window, and a quality-control reviewer checked each annotation, yielding 1,144 overview verdicts and 4,672 window verdicts. Reference-overlap metrics do not separate the captions reviewers accepted from those they rejected: BLEU-4 reaches an AUC of 0.510 and a CIDEr-style tf--idf cosine 0.528, and restricting references to reviewer corrections leaves BLEU-4 at 0.479 (95\% CI 0.431--0.529), indistinguishable from chance. Review granularity changes what is detected. The same annotators record camera and shot errors on 2.7\% of captions when reading the overview and on 12.8\% when reading window by window, a 4.7$\times$ gap that holds across every robustness check we ran, on the mode most likely to need a near-total rewrite. Correct-overview rates span 54.1--72.7\%, but only 4 of 15 pairwise gaps survive Bonferroni correction (5 after adjusting for reviewer and clip effects), and accuracy and derived severity order the models differently. Models with statistically indistinguishable accuracy have sharply different failure profiles, which no single scalar shows. We argue that caption evaluation should record what was wrong rather than only how often, and we report the taxonomy, the per-model profiles, the corpus properties that predict difficulty, and the code that regenerates every number in the paper.
\end{abstract}

\section{Introduction}

A caption used to be metadata: alternate text, or a keyword for search. It is now the primary artefact in a growing set of systems. Retrieval, clipping and search over video libraries consume the caption itself, which means a description that reads well but places an action three seconds from where it occurred is a defect rather than a stylistic preference.

That shift leaves two questions for anyone selecting a captioner: which model suits a given corpus, and when a model is wrong, what specifically is wrong with it. The instruments built for the first question were designed for a different task, and few address the second.

This paper reports CaptionBench, our evaluation of six video captioning models on a shared corpus, in which trained annotators did not grade captions but post-edited them.\footnote{A non-technical summary of this study was published as ``Introducing CaptionBench'', \url{https://www.deccan.ai/research/introducing-caption-bench-video-captioning-benchmark}. Where the two differ, the figures in this paper supersede the summary.} Every edit is typed against a taxonomy fixed before annotation. The approach is not new in kind: human-targeted edit rates \citep{snover2006ter} and typed error annotation \citep{lommel2014mqm,freitag2021experts} are standard in machine translation, and error typologies are used to audit summarisation metrics \citep{pagnoni2021frank}. Our contribution is to apply typed post-editing to dense video description, where the caption has both a whole-clip summary and time-windowed parts, and to measure what that instrument shows that a scalar does not. Three findings follow.
\begin{enumerate}
\item \textbf{Overlap metrics do not recover the human verdict on this task.} BLEU-4 and a CIDEr-style cosine separate accepted from rejected overviews at chance, with or without reviewer-written references, and no better than caption length alone (\secref{sec:metrics}). That such metrics track human judgement poorly is long established for image captions \citep{elliott2014comparing,kilickaya2017reevaluating} and for video \citep{graham2018direct}; we confirm it with intervals and controls on long, multi-part descriptions.
\item \textbf{The granularity of the read changes what is detected.} Holding annotator, footage and taxonomy constant and varying only the unit of the read changes recorded camera-error incidence by 4.7$\times$ (\secref{sec:granularity}). Camera weakness in video-language models is documented elsewhere \citep{lin2025camerabench,liu2025shotbench,kong2025tuna}; the result here concerns the evaluation instrument, not the models.
\item \textbf{Scalar rankings conceal actionable structure.} Models whose accuracy cannot be statistically separated differ sharply in \emph{what} they get wrong, and a model second on accuracy is last on derived severity (\secref{sec:profiles}).
\end{enumerate}
We also report hypotheses that did not survive (\secref{sec:negative}), including a collider artefact that will recur in any analysis of failure labels recorded only on rejected units.

\section{Related work}
\label{sec:related}

\paragraph{Caption metrics and human judgement.} Reference-overlap metrics, namely BLEU \citep{papineni2002bleu}, METEOR \citep{banerjee2005meteor} and CIDEr \citep{vedantam2015cider}, were built for short outputs with a narrow space of correct answers, and their agreement with human judgement on image captions is weak \citep{elliott2014comparing,kilickaya2017reevaluating}. TRECVID's direct-assessment study reached the same conclusion for video \citep{graham2018direct}. Reference-free scores compare a caption with frames \citep{hessel2021clipscore,shi2022emscore}. Large language models are increasingly used as judges \citep{chan2023clair,tong2025gveval}, with documented biases \citep{chen2024mllmjudge}.

\paragraph{Detailed and dense video description.} Dense captioning localises descriptions in time \citep{krishna2017dense}, with SODA evaluating the resulting sequences \citep{fujita2020soda}. Recent benchmarks score long descriptions by decomposing them into checkable units or questions. VDC and VDCscore \citep{chai2025auroracap} do this for detailed captions. DREAM-1K and AutoDQ \citep{wang2024tarsier} score event-level precision and recall. CAPability \citep{liu2025capability} separates correctness from thoroughness. TUNA \citep{kong2025tuna} targets temporal and camera elements. Shot2Story \citep{han2025shot2story} and Vript \citep{yang2024vript} pair shot-level descriptions with a summary, as the captions here do. Hallucination benchmarks ask whether asserted content exists: CHAIR \citep{rohrbach2018object}, POPE \citep{li2023pope}, FaithScore \citep{jing2024faithscore} and VideoHallucer \citep{wang2024videohallucer}. ARGUS \citep{rawal2025argus} scores free-form video captions for hallucination and omission separately, which corresponds to two of our modes.

\paragraph{Camera and shot understanding.} CameraBench \citep{lin2025camerabench} and ShotBench \citep{liu2025shotbench} show that vision-language models misread camera motion and shot grammar. We do not claim to discover this weakness. Our finding is that an overview-level read largely fails to record it.

\paragraph{Post-editing and error typologies.} HTER measures quality as the edits a human makes to reach an acceptable output \citep{snover2006ter}. MQM types each error and assigns it a severity \citep{lommel2014mqm}, and expert MQM annotation changes metric rankings relative to crowd ratings \citep{freitag2021experts}. FRANK uses a typology to ask which errors summarisation metrics detect \citep{pagnoni2021frank}. ImageInWords builds hyper-detailed image descriptions by human correction of model drafts \citep{garg2024imageinwords}. \citet{vanmiltenburg2020gradations} show that edits of similar size can differ greatly in severity, which bears directly on our derived severity measure (\secref{sec:limitations}).

\section{Corpus, models and protocol}
\label{sec:protocol}

\paragraph{Corpus.} The corpus has 196 short clips, with median duration 10.0\,s (range 4.8--170.5\,s) and 181 under thirty seconds. 107 are Pexels stock clips, 15 of them trimmed or altered. 35 are first-person household-activity clips. The remaining 54 are internal uploads whose original source was not recorded. Twelve files are manipulated versions of source footage: 9 are time-reversed, one is frame-shuffled, one colour-banded and one datamoshed. We keep them in the main analysis and report every headline result without them (\tabref{tab:robust}, \secref{sec:limitations}).

\paragraph{Clip metadata.} Each clip carries labels for content type, camera platform, shot scale, setting, scene complexity and motion intensity. These were inferred before captioning, on 15 August 2026, by a text-only language model (Gemini 3.1 Flash-Lite, temperature 0) from earlier human-written descriptions of each clip; the model did not see the video. Content type is a fixed mapping of the inferred domain. We use the labels only for the corpus analysis in \secref{sec:difficulty}. They are model-inferred and not independent of the evaluated systems' training lineage: one evaluated model is a Gemini sibling.

\paragraph{Models.} Six systems captioned every clip with the same prompt template at temperature~0: Qwen3.8-Max, Muse Spark 1.2, Kimi K3, Seed 2.1 Turbo, Gemini 3.1 Pro and Qwen3.8-27B. Five developers are represented. Qwen3.8-27B is the open dense member of the Qwen3.8 series and Qwen3.8-Max its proprietary flagship. Audio was stripped before inference. Inputs were not otherwise identical across models (\tabref{tab:runs}, Appendix~\ref{app:runs}):
\begin{itemize}
\item frame sampling and resolution were left to each provider, giving a 12-fold spread in input tokens per second of video;
\item the two Qwen3.8 models received re-encoded copies of oversized clips;
\item outputs that failed schema validation were retried with the validator's message appended (Kimi K3 on 55 clips, Qwen3.8-Max 38, Qwen3.8-27B 22, Seed 2.1 Turbo 7);
\item five Qwen3.8-Max outputs that never passed validation were reviewed as returned.
\end{itemize}
A seventh system, Qwen3.5-397B, was captioned and reviewed but is excluded. A provider defect forced 86 of its clips to be regenerated, and the two runs (63.7\% and 70.6\% correct, on different clips) do not describe one system.

\paragraph{Output structure.} Each model returns, per clip, an \emph{overview}: a single paragraph describing the whole video, with median length 64--95 words depending on the model. It also returns a \emph{timeline}: 3.7--4.5 time-bounded windows on average, each with a shot description (camera behaviour) and a scene description (what occurs in front of the camera), totalling a median of 224 words. A one-line visual-style note and a language field were also returned. The style note was reviewed only for checkable factual claims, and the language was judged wrong on 3 of 1,144 captions. Overview and timeline are written independently: the median character-level similarity between a model's overview and its own first window is 0.067, and the maximum in the corpus is 0.63.

\paragraph{Annotation and quality control.} Annotators watched each clip and post-edited the model's caption against a written rubric fixed before annotation, correcting every material error and preserving accurate wording. Each caption received two judgements:
\begin{itemize}
\item \textbf{Overview} --- accept the overview as written, or reject it, tag the applicable failure modes and write a corrected version.
\item \textbf{Windows} --- accept or reject each timeline window on its own terms, with modes from the same taxonomy, plus three segmentation questions answered once per caption.
\end{itemize}
The taxonomy has five content modes (factual error, important omission, camera or shot misunderstanding, unsupported inference, temporal or order error), a segmentation mode, and a free-text residual; \tabref{tab:cost} gives the definition annotators were shown for each. The rubric instructs annotators to tag a factual error caused by a camera misreading as camera only. The data do not always follow this rule: 13 of the 31 overviews and 92 of the 305 windows tagged camera are also tagged factual. We report labels as recorded and show that enforcing the rule changes no conclusion (\tabref{tab:robust}).

Every annotation then passed a quality-control (QC) reviewer, who graded it and could edit it. The QC pass is not an independent second annotation. It saw the annotator's work, and in 1,027 of 1,143 captions every overview field is unchanged. QC changed the overview verdict on 20 of 1,143 captions (16 accept to reject, 4 reject to accept; Cohen's $\kappa$ between the two passes, 0.961, measures that edit rate and not inter-annotator agreement). It changed 119 of 4,688 window verdicts, and window rejection moved from 29.1\% to 29.9\%. We use the QC-pass overview verdict as the final verdict and the annotator-pass window verdicts. Keeping the annotator pass at the window level holds the person constant for the granularity comparison, and \tabref{tab:robust} shows the alternatives give the same result. We did not collect independent double annotation, so inter-annotator agreement is not measured (\secref{sec:limitations}).

\paragraph{Counts.} 1,176 captions were planned (196 clips $\times$ 6 models). 1,154 were annotated, and 1,144 carry a valid final overview verdict, 752 accepted and 392 rejected; 10 were dropped for a blank or corrupted verdict field. These 1,144 captions carry 4,688 timeline windows, of which 4,672 received a verdict. Twenty-one annotators took part. Eighteen reviewed at least 30 captions, and two reviewed 200 and 192, together 34\% of the data. Annotation ran from 19 to 28 August 2026. Qwen3.8-27B was added last and annotated on 27--28 August by 16 annotators, all of whom also reviewed other models.

\paragraph{Samples.} Model comparisons of accuracy use the 172 clips on which all six models carry a final verdict, so every model contributes $n=172$ (1,032 captions). Descriptive profiles, such as mode incidence, component survival and repair direction, use all 1,144 captions ($n=187$--194 per model) and say so in their captions. Appendix~\ref{app:paired} repeats the main per-model tables on the paired subset. The orderings discussed in the text hold there, except that Seed 2.1 Turbo edges past Gemini 3.1 Pro on segmentation (18.0\% against 16.9\%).

\paragraph{Statistics.} Rates carry Clopper--Pearson exact intervals \citep{clopper1934}. Model gaps use a paired bootstrap over clips \citep{efron1993bootstrap} with 20,000 resamples, with Bonferroni correction across all 15 pairs applied once \citep{dror2018hitchhiker}. Reviewer- and clip-adjusted gaps come from an additive model (model + annotator + clip) refitted inside each of 4,000 clip-bootstrap resamples, so the intervals carry the uncertainty of the adjustment. Window-level and AUC intervals use the clip bootstrap, because captions on the same clip share footage and references. Properties measured per clip are tested with clips as the unit. Failure modes are recorded only on rejected units, so every incidence figure is a floor.

\section{Do overlap metrics recover human judgement?}
\label{sec:metrics}

Three families of instrument are available, and each answers a question adjacent to the one a caption buyer is asking.

\emph{Reference-overlap metrics} \citep{papineni2002bleu,banerjee2005meteor,vedantam2015cider} score $n$-gram agreement against a reference. The overviews here run 64--95 words, the timelines over 200 more, and there are many defensible ways to describe ten seconds of footage, so overlap penalises valid paraphrase and rewards generic text. References also cost roughly what a correction costs to produce, while carrying strictly less information: a correction additionally identifies what the model got wrong.

\emph{Embedding scores} such as CLIPScore \citep{hessel2021clipscore} and EMScore \citep{shi2022emscore} compare a caption with frames. A caption naming the right objects in the wrong order, or attributing a camera movement to the subject, is largely unpunished, because the frames contain everything mentioned. Such errors are common in this corpus. Segmentation failure runs 6.8--18.1\% across models, and camera errors are the mode most likely to need a near-total rewrite (\tabref{tab:cost}).

\emph{Hallucination benchmarks} \citep{rohrbach2018object,li2023pope,jing2024faithscore,wang2024videohallucer,rawal2025argus} ask whether asserted content exists. That is the right question for factual errors, but these benchmarks do not cover camera, segmentation or temporal failures. A judge that scores only the summary paragraph also inherits the blind spot documented in \secref{sec:granularity}.

\paragraph{Design.} Every clip carries up to six reviewer-approved descriptions: accepted overviews as written, and reviewer corrections where the overview was rejected. We score each candidate overview against the approved texts of the \emph{other} models on the same clip and ask whether the score separates accepted from rejected candidates; the readout is the AUC, where 0.5 carries no information. Two reference conditions are reported: \emph{approved}, using all approved text, and \emph{gold}, restricted to reviewer corrections.
\begin{itemize}
\item \textbf{BLEU-4} is multi-reference sentence BLEU, unsmoothed, on lower-cased word tokens.
\item \textbf{CIDEr-style} is a tf--idf $n$-gram cosine ($n=1$--4) averaged over references and $n$, scaled by 10. It has no clipping or length penalty, so it is not CIDEr-D.
\end{itemize}
One accepted caption, on a clip with no other verdict, has no reference and drops out (751 accepted, 392 rejected). In the gold condition, an accepted candidate on a clip with $k$ rejections receives $k$ references and a rejected candidate $k-1$, which lowers the scores of rejected candidates for reasons unrelated to quality. The \emph{matched} column of \tabref{tab:metrics} removes this artefact. It compares accepted and rejected candidates on the same clip against an identical reference set. As controls, we score overview length alone, and a video-grounded LLM judge that we had run for a separate purpose: Gemini 3.1 Pro listed the material defects in each original caption, before seeing the annotation.

\begin{table}[t]
\centering
\caption{Agreement between automatic scores and the human accept/reject verdict on the overview, over all captions with a final verdict. AUC 0.5 means the score carries no information; intervals are 95\% bootstrap over clips. \emph{Matched}: accepted and rejected candidates on the same clip scored against an identical reference set. $^\dagger$Gemini 3.1 Pro judge, 648 accepted / 301 rejected captions from five models (Qwen3.8-27B was not judged). The judge counts defects in the overview and timeline together, and Gemini 3.1 Pro is itself a candidate.}
\label{tab:metrics}
\tablebox{\begin{tabular}{llccccc}
\toprule
References & Score & Accepted & Rejected & AUC & 95\% CI & Matched \\
\midrule
Approved ($n$=751/392) & BLEU-4 & 0.358 & 0.355 & 0.510 & [0.473, 0.546] & 0.520 \\
 & CIDEr-style cosine & 1.834 & 1.782 & 0.528 & [0.488, 0.567] & 0.526 \\
\addlinespace
Gold ($n$=579/352) & BLEU-4 & 0.206 & 0.210 & 0.479 & [0.431, 0.529] & 0.500 \\
 & CIDEr-style cosine & 1.768 & 1.729 & 0.518 & [0.475, 0.562] & 0.524 \\
\addlinespace
Controls & Overview length (shorter $\to$ accept) & 85w & 86w & 0.524 & [0.486, 0.562] & -- \\
& Video-LLM judge, fewer defects$^\dagger$ & & & 0.558 & [0.519, 0.597] & -- \\
\bottomrule
\end{tabular}
}
\end{table}

\paragraph{Result.} \tabref{tab:metrics} shows no usable agreement. BLEU-4 reaches 0.510 and the CIDEr-style cosine 0.528. With reviewer corrections as references BLEU-4 is 0.479, and its interval [0.431, 0.529] includes 0.5; rejected candidates even score marginally higher than accepted ones on average (0.210 against 0.206), a difference well within noise. On the 931 candidates scored under both conditions, the gold--approved difference is $-0.025$ [$-0.075$, 0.025], so restricting references to corrections does not measurably change the result. With matched references, all four AUCs lie between 0.500 and 0.526, and length alone reaches 0.524 (accepted overviews have a median of 85 words, rejected 86).

The gold references are also less independent of the candidates than they appear: a reviewer correction keeps a median 93.5\% of its own draft's tokens (interquartile range 84.0--98.1\%). Of the automatic scores examined, only the video-grounded LLM judge excludes chance, at 0.558 [0.519, 0.597]. That is a weak signal from a judge with known problems (\secref{sec:negative}).

At model level, mean BLEU-4 and the paired correct-overview rate correlate at Spearman $\rho=+0.60$. With six models that is weak evidence (exact two-sided $p=0.24$), and the ordering it implies is wrong where it matters. Qwen3.8-27B ranks third of six on mean BLEU-4 at 0.368, above Seed 2.1 Turbo, Kimi K3 and Gemini 3.1 Pro, while reviewers place it last at 54.1\%. Part of its BLEU advantage comes from its sibling Qwen3.8-Max being among its references. Scoring every model against three non-Qwen references moves it to fourth (0.277), still above Kimi K3 and Gemini 3.1 Pro.

\subsection{Does the ordering survive a stricter bar?}

\begin{figure}[t]
\centering
\includegraphics[width=0.92\linewidth]{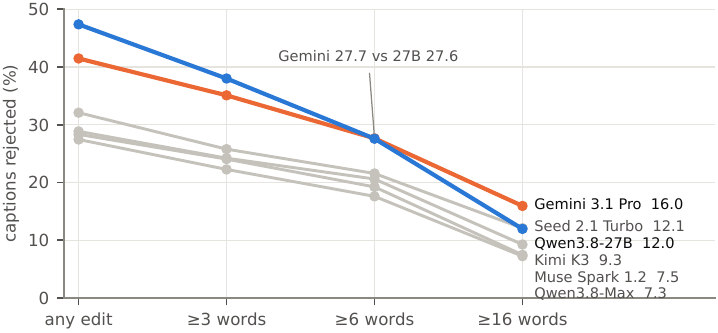}
\caption{Overview intervention rate recomputed while progressively discarding small corrections (\tabref{tab:threshold}). The four frontier models in grey hold their order throughout. Gemini 3.1 Pro and Qwen3.8-27B tie at the six-word bar and swap places beyond it: much of Qwen3.8-27B's excess comes from small corrections.}
\label{fig:threshold}
\end{figure}

A rejection rate is only meaningful if the rejections are substantive. Raising the evidence threshold, counting a caption as failed only when the correction changed at least $k$ words, tests whether the model ordering is an artefact of trivial edits (\tabref{tab:threshold}, \figref{fig:threshold}). Only 3 of the 392 rejections change no word at all, and 34.7\% change fewer than six.

\begin{table}[t]
\centering
\caption{Overview intervention rate at increasing evidence thresholds, over all captions with a final verdict. A caption counts as failed only if the correction changed at least that many words (word-level diff, replacements counted at the longer side). Appendix~\ref{app:paired} gives the same table on the 172-clip paired subset.}
\label{tab:threshold}
\tablebox{\begin{tabular}{lccccc}
\toprule
Model & $n$ & Any correction & $\geq$3 words & $\geq$6 words & $\geq$16 words \\
\midrule
Qwen3.8-Max & 193 & 27.5 & 22.3 & 17.6 & 7.3 \\
Muse Spark 1.2 & 187 & 28.3 & 24.1 & 19.3 & 7.5 \\
Kimi K3 & 194 & 28.9 & 24.2 & 20.6 & 9.3 \\
Seed 2.1 Turbo & 190 & 32.1 & 25.8 & 21.6 & 12.1 \\
Gemini 3.1 Pro & 188 & 41.5 & 35.1 & 27.7 & 16.0 \\
Qwen3.8-27B & 192 & 47.4 & 38.0 & 27.6 & 12.0 \\
\bottomrule
\end{tabular}
}
\end{table}

Gemini 3.1 Pro stays above the four leading models at every threshold (27.7\% at the six-word bar, against Qwen3.8-Max's 17.6\%). Qwen3.8-27B does so for the first three columns and falls level with Seed 2.1 Turbo at the fourth. The ordering of the four leading models is stable across all four columns. The two weakest models trade places: Qwen3.8-27B leads with any correction (47.4\% against 41.5\%), they tie at six words (27.6\% and 27.7\%), and at sixteen words Qwen3.8-27B (12.0\%) falls below Gemini 3.1 Pro (16.0\%) and level with Seed 2.1 Turbo (12.1\%). Qwen3.8-27B's deficit is therefore concentrated in many small fixes, and Gemini 3.1 Pro's in larger rewrites, consistent with the repair-direction results in \secref{sec:corrections}.

\section{Review granularity changes what is detected}
\label{sec:granularity}

\begin{figure}[t]
\centering
\includegraphics[width=0.92\linewidth]{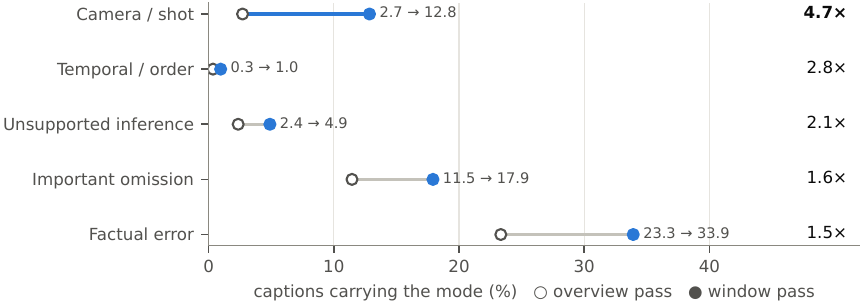}
\caption{The data of \tabref{tab:granularity}. Camera and shot is the outlier in both position and ratio.}
\label{fig:granularity}
\end{figure}

Each annotator judged the same output twice against the same taxonomy, once as a whole-clip overview and once window by window. The comparison therefore holds annotator, footage and taxonomy constant and varies only the unit of the read (\tabref{tab:granularity}, \figref{fig:granularity}).

\begin{table}[t]
\centering
\caption{Share of captions recorded as carrying each mode, by review granularity, over all 1,144 captions. Overview pass: modes tagged on the rejected overview. Window pass: modes tagged on any rejected window of the same caption. $k$ is the number of captions.}
\label{tab:granularity}
\tablebox{\begin{tabular}{lccc}
\toprule
Failure mode & Overview pass, \% ($k$) & Window pass, \% ($k$) & Ratio \\
\midrule
Camera / shot & 2.7 (31) & 12.8 (147) & \textbf{4.7$\times$} \\
Temporal / order & 0.3 (4) & 1.0 (11) & 2.8$\times$ \\
Unsupported inference & 2.4 (27) & 4.9 (56) & 2.1$\times$ \\
Important omission & 11.5 (131) & 17.9 (205) & 1.6$\times$ \\
Factual error & 23.3 (267) & 33.9 (388) & 1.5$\times$ \\
\bottomrule
\end{tabular}
}
\end{table}

Every mode is recorded more often under the closer read, which is expected. Camera and shot breaks the pattern at 4.7$\times$, 1.7 times the next-largest ratio (temporal, 2.8$\times$). A camera error is camera behaviour read as world behaviour: a pan described as the subject moving, or a dolly described as objects drifting. The gap does not depend on our analysis choices (\tabref{tab:robust}). It holds when both levels come from the annotator's own pass (4.9$\times$) and when both come from the QC pass (4.7$\times$). It is unchanged when the camera-over-factual rule is enforced, and it holds on the paired subset, without the manipulated clips, and with the seventh model included.

\begin{table}[t]
\centering
\caption{Ratio of window-pass to overview-pass incidence for camera and shot errors under alternative analysis choices.}
\label{tab:robust}
\tablebox{\begin{tabular}{lc}
\toprule
Variant & Camera / shot ratio \\
\midrule
Primary: QC-pass overview, annotator-pass windows, 1,144 captions & 4.7$\times$ \\
Annotator pass at both levels (same person, same sitting) & 4.9$\times$ \\
QC pass at both levels & 4.7$\times$ \\
Camera-over-factual precedence enforced on every unit & 4.7$\times$ \\
172-clip paired subset & 5.5$\times$ \\
Excluding the 12 manipulated clips & 4.5$\times$ \\
Seven models (Qwen3.5-397B included) & 4.6$\times$ \\
\bottomrule
\end{tabular}
}
\end{table}

Two mechanisms are available, and this design cannot fully separate them. The window form carries an explicit shot field, so it elicits a camera claim the overview never had to make: an unmade claim cannot be marked wrong. And an annotator assessing a whole paragraph may accept an error that does not disturb the overall reading. Only the second generalises: no other mode has a dedicated field, and the remaining modes gain 1.5--2.8$\times$ regardless. Our reading is that the form accounts for camera's excess over the other modes, while the unit of review accounts for the effect common to all of them. An explicit camera question at overview level would separate the two.

The practical consequence is that any evaluation reading only the summary, human or automated, systematically under-counts the mode most likely to require a near-total rewrite. When a camera error is caught, 21.6\% of the window repairs it triggers are near-total rewrites, against 5.5\% for omissions (\tabref{tab:cost}).

\subsection{A worked example: one misread, propagated}

The mechanism behind both the granularity gap and the clustering result below is visible in a single record: a 9.4-second product clip captioned by Gemini 3.1 Pro, rejected by the annotator, with the rejection upheld unchanged by QC.

\begin{quote}\small
\textbf{Overview as generated.} ``Small, dark, spherical objects are scattered on a white surface. They roll across the surface from left to right and \emph{towards the foreground}. \emph{The camera pans right} to follow their movement. Eventually, most of the objects roll out of the frame, leaving only a few behind.''

\textbf{Correction.} ``Small, dark, spherical objects are scattered on a white surface. They \emph{jitter, bounce, and roll} across the surface from left to right and \emph{towards the background as the camera pans to follow} their movement. Eventually, most of the objects roll out of the frame, leaving only a few behind.''
\end{quote}

The model made one underlying error: it read a forward dolly first as a stationary camera and then as a rightward pan. Because the camera was believed static, the objects' apparent outward drift had to be them rolling \emph{towards the foreground}, so one camera misattribution became a factual error about the scene in every window that depended on it. The annotator tagged both camera and factual error at overview level and in all three windows, contrary to the precedence rule; this is the co-tagging noted in \secref{sec:protocol}. At overview level it is one rejection. At window level it is three rejected windows, and the shot field forces the camera claim into the open in each of them. The same record shows why the two levels are scored separately, why failures cluster within a caption, and why this model's repairs lengthen the text: the correction must \emph{add} what the original omitted.

\subsection{An accepted overview is not a clean caption}

\begin{figure}[t]
\centering
\includegraphics[width=0.92\linewidth]{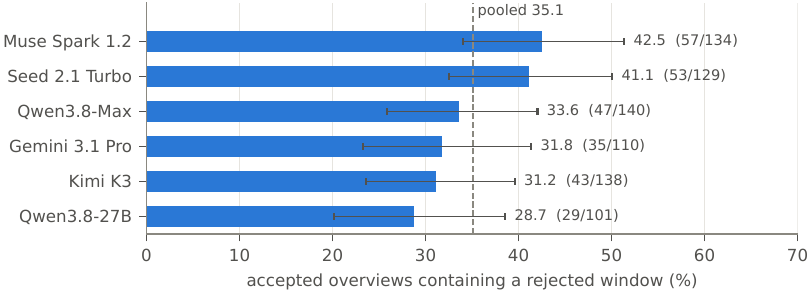}
\caption{Share of overviews accepted whole that contain at least one window the same annotator rejected, with 95\% Clopper--Pearson intervals and counts, all captions. The dashed line is the pooled rate.}
\label{fig:hidden}
\end{figure}

Among overviews an annotator accepted whole, the share with at least one window the same annotator rejected is 42.5\% for Muse Spark 1.2, 41.1\% for Seed 2.1 Turbo, 33.6\% for Qwen3.8-Max, 31.8\% for Gemini 3.1 Pro, 31.2\% for Kimi K3 and 28.7\% for Qwen3.8-27B, 35.1\% pooled (\figref{fig:hidden}). Muse Spark 1.2 therefore hides a rejected window in one of every 2.4 overviews accepted whole. Conversely, 12.5\% of rejected overviews (49 of 391) sit above a timeline with no rejected window. Even the cleanest model on this measure conceals a window failure in more than a quarter of the captions that passed at overview level. This is the second measure, after severity (\secref{sec:profiles}), on which Muse Spark 1.2 looks clean at the top level and does not hold up underneath.

\subsection{Rejections cluster within a caption}

\begin{figure}[t]
\centering
\includegraphics[width=0.92\linewidth]{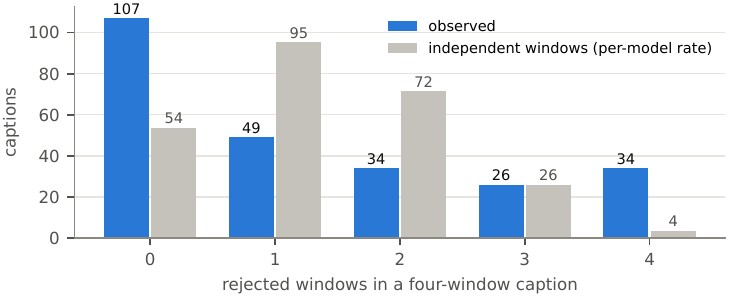}
\caption{Rejected windows per caption among the 250 captions with exactly four windows, against what independent windows at each model's own rejection rate would give. The middle is hollow and both tails are heavy.}
\label{fig:clustering}
\end{figure}

Among the 250 captions with exactly four windows, the observed rate of rejected windows is 33.1\%. If windows failed independently at each model's own rate, we would expect 54 captions with no rejected window and 4 with all four rejected. We observe 107 and 34 (\figref{fig:clustering}; $\chi^2=341$, 3 d.f., $p<10^{-70}$). Captions are mostly right or mostly wrong, and rarely partly wrong. This is consistent with a single misread propagating through every window that depends on it, as in the worked example. It is equally consistent with clip-level difficulty and with annotator strictness, both of which correlate window verdicts within a caption, and this design cannot apportion the effect among the three.

The segmentation questions show an asymmetry. Models insert a boundary where nothing changed on 7.4\% of captions and merge two distinct events on 3.8\%, so they impose structure on continuous footage about twice as often as they miss structure that is present. A further 3.1\% of captions have a misplaced boundary.

\section{Accuracy, severity and failure profiles}
\label{sec:profiles}

\begin{figure}[t]
\centering
\includegraphics[width=0.92\linewidth]{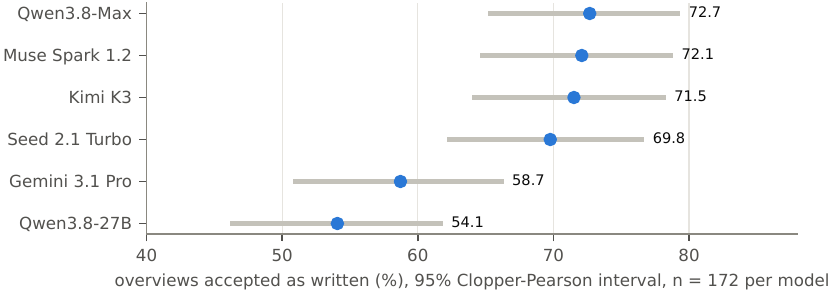}
\caption{Correct-overview rate on the 172-clip paired subset with 95\% Clopper--Pearson intervals (\tabref{tab:primary}). The four leading models overlap almost entirely, and so do the two trailing ones.}
\label{fig:accuracy}
\end{figure}

\paragraph{Accuracy.} The correct-overview rate is the share of overviews accepted exactly as written. On the paired subset it spans 54.1\% to 72.7\% (\tabref{tab:primary}), with intervals of about $\pm$7 points. Only \textbf{4 of 15 pairwise gaps survive Bonferroni correction}, all four separating Qwen3.8-27B from the four leading models (Appendix~\ref{app:pairwise}). After adjustment for annotator and clip effects, a fifth pair separates, Muse Spark 1.2 against Gemini 3.1 Pro (14.1 points, Bonferroni interval 0.8--27.8). The four models between 69.8\% and 72.7\% are not distinguishable from one another under either analysis: the leader is inside the tie, not above it. Neither are Gemini 3.1 Pro and Qwen3.8-27B, which sit 4.6 points apart (58.7\% against 54.1\%; \figref{fig:accuracy}). The leaderboard is shorter than it looks.

 The window-level correct rate orders the models similarly (Spearman $\rho=0.83$ across the six, exact $p=0.058$).

\begin{table}[t]
\centering
\caption{Primary measures on the 172-clip paired subset ($n=172$ captions per model). Correct overview: accepted as written (Clopper--Pearson interval). Correct window: share of windows accepted (bootstrap over clips). Critical: share of captions with at least one rejected window whose corrected text has character-level similarity below 0.55 to the original (Clopper--Pearson). Critical is derived from edit size, not recorded by reviewers (\secref{sec:limitations}).}
\label{tab:primary}
\tablebox{\begin{tabular}{lcccccc}
\toprule
& \multicolumn{2}{c}{Correct overview} & \multicolumn{2}{c}{Correct window} & \multicolumn{2}{c}{Critical (derived)} \\
\cmidrule(lr){2-3}\cmidrule(lr){4-5}\cmidrule(lr){6-7}
Model & \% & 95\% CI & \% & 95\% CI & \% & 95\% CI \\
\midrule
Qwen3.8-Max & 72.7 & [65.4, 79.2] & 80.7 & [76.7, 84.5] & 5.8 & [2.8, 10.4] \\
Muse Spark 1.2 & 72.1 & [64.8, 78.7] & 68.4 & [62.2, 74.1] & \textbf{16.3} & [11.1, 22.7] \\
Kimi K3 & 71.5 & [64.1, 78.1] & 73.7 & [68.1, 78.8] & 9.9 & [5.9, 15.4] \\
Seed 2.1 Turbo & 69.8 & [62.3, 76.5] & 71.6 & [65.6, 77.3] & 10.5 & [6.3, 16.0] \\
Gemini 3.1 Pro & 58.7 & [51.0, 66.2] & 65.0 & [58.8, 71.1] & 8.7 & [5.0, 14.0] \\
Qwen3.8-27B & 54.1 & [46.3, 61.7] & 64.3 & [58.4, 70.2] & 14.0 & [9.1, 20.0] \\
\bottomrule
\end{tabular}
}
\end{table}

\paragraph{Severity reorders the field.} We call a caption \emph{critical} when at least one of its rejected windows needed a near-total rewrite: the corrected window text has character-level similarity below 0.55 to the original. Muse Spark 1.2 is second on accuracy and last on this measure, at 16.3\% against Qwen3.8-Max's 5.8\%. Its severity distribution is bimodal: it has the lowest share of captions with a moderate window rewrite (similarity 0.55--0.85; 11.0\%) and the highest share with a near-total one. When Muse Spark 1.2 gets a caption right it is usually clean; when it gets one wrong it usually has to be redone. Selection on accuracy alone and selection on severity alone yield different recommendations.

\paragraph{What a near-total rewrite looks like.} A 10.6-second domestic clip captioned by Gemini 3.1 Pro was rejected by the annotator, with the rejection upheld unchanged by QC (QC grade 5 of 5). It is tagged factual error at overview level and in all three of its windows, which were all rejected:
\begin{quote}\small
\textbf{Overview as generated.} ``First, \emph{two people} are seen from the arms down, sitting on a white marble floor. \emph{The person on the left} holds a dark, patterned fabric, while \emph{the person on the right} holds a green plastic bottle. Then, the person on the right moves the bottle out of view and uses their hands to help manipulate the fabric. \emph{Both individuals squeeze and twist the fabric together.} Finally, the camera shifts downwards, focusing on the hands and the floor.''

\textbf{Correction.} ``\emph{A person is seen wiping a tiled wall} with both arms down, sitting on a white marble floor. \emph{The left hand} holds a dark, patterned fabric, while \emph{the right} holds a green plastic bottle. \emph{The person sprinkles some liquid from the bottle over the cloth and then cleans the tiled wall with it.} Then, the right hand moves the bottle out of view. Finally, the camera shifts downwards, focusing on the hands and the floor.''
\end{quote}
Two hands became two people, and the model then built an interaction between them that never happened, while missing what the person was actually doing. The correction changed 58 of the overview's 77 words (similarity 0.49), and the caption counts as critical. This is the kind of repair the critical rate counts; the pooled rates matter more than any single record.

\paragraph{Frequency is not cost.} \tabref{tab:cost} gives the definition, incidence and severity of each mode. Ordered by frequency, the content modes run factual error, omission, camera. Ordered by how many captions per hundred carry a near-total rewrite attributable to the mode, camera (4.1) moves ahead of omission (1.6). Camera is the least frequent of the three and the one most likely to need a near-total rewrite when it occurs: 21.6\% of camera window repairs are near-total, against 16.0\% for factual errors and 5.5\% for omissions, and 32.0\% of captions carrying a camera error have at least one such window.

\begin{table}[!htbp]
\centering
\caption{The failure-mode taxonomy, frozen before annotation, with each mode's incidence and severity pooled across the six models (window detector, all 1,144 captions). Definitions are those shown to annotators; the decision tests are our one-line summaries and were not part of the rubric. \emph{Captions}: share of captions carrying the mode at least once. \emph{Near-total}: share of that mode's window repairs rewritten to character-level similarity below 0.55. \emph{Per 100}: captions per hundred with at least one near-total window repair carrying the mode. Segmentation is judged once per caption and has no window-level severity. All figures are floors, since modes were recorded only on rejected units.}
\label{tab:cost}
\tablebox{\begin{tabular}{@{}>{\raggedright\arraybackslash}p{2.3cm}>{\raggedright\arraybackslash}p{7.2cm}ccc@{}}
\toprule
Mode & Definition given to annotators \newline \emph{Decision test (our summary)} & Captions (\%) & Near-total (\%) & Per 100 \\
\midrule
Factual error & An observable fact is stated incorrectly: wrong person, object, action, attribute, count, location or spatial relation. \newline \emph{Can a reviewer point at a frame and say ``not that''?} & 33.9 & 16.0 & 7.5 \\
\addlinespace
Important omission & A central event or piece of information needed to describe the content adequately is missing; optional background detail is not marked. \newline \emph{Would a reader holding only the caption be misled?} & 17.9 & 5.5 & 1.6 \\
\addlinespace
Camera / shot & Camera motion, zoom, cuts, framing, occlusion or reflection is mistaken for something happening in the physical scene. \newline \emph{Did the pixels move because the camera moved, or the world?} & 12.8 & 21.6 & 4.1 \\
\addlinespace
Segmentation & Windows merge distinct events, miss a shot change, are placed on a grid rather than on content, or are timestamped materially away from the event. \newline \emph{Are the timestamps where the events actually are?} & 13.4 & -- & -- \\
\addlinespace
Unsupported inference & Intent, emotion, identity, cause or relationship is asserted where the video does not support it, including identification from outside knowledge. \newline \emph{Could any viewer verify this from the footage alone?} & 4.9 & 19.0 & 1.2 \\
\addlinespace
Temporal / order & Event order, timing, duration, repetition or simultaneity is materially misrepresented. \newline \emph{Is what happened right but when wrong?} & 1.0 & 0.0 & 0.0 \\
\bottomrule
\end{tabular}
}
\end{table}

\begin{figure}[t]
\centering
\includegraphics[width=0.92\linewidth]{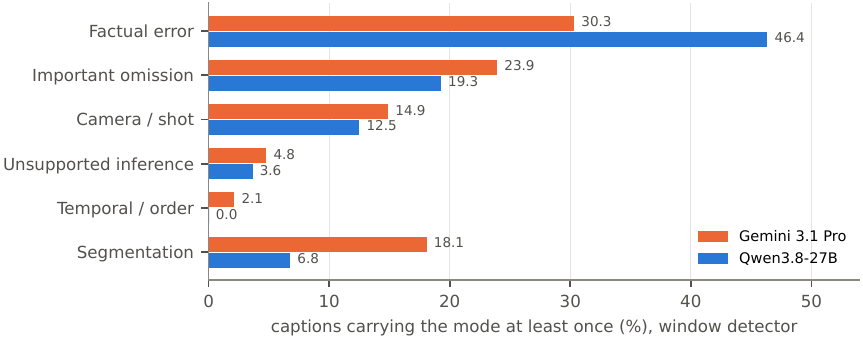}
\caption{Gemini 3.1 Pro against Qwen3.8-27B, two models 4.6 points apart on accuracy that cannot be statistically separated: share of captions carrying each mode at least once, window detector, all captions. Qwen3.8-27B is far worse on factual error and far better on segmentation; Gemini 3.1 Pro is the reverse on both and also worse on omission, camera and temporal order.}
\label{fig:pair}
\end{figure}

\begin{figure}[t]
\centering
\includegraphics[width=\linewidth]{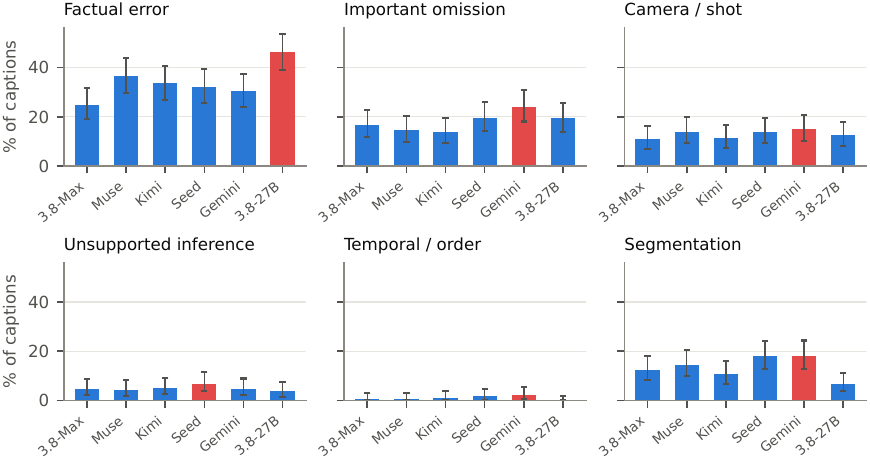}
\caption{Failure-mode incidence by model with 95\% Clopper--Pearson intervals, one panel per mode, models ordered by paired accuracy, the highest bar in each panel in red. Camera is flat across the field; factual error and segmentation are not.}
\label{fig:profiles}
\end{figure}

\paragraph{Profiles diverge where scores do not.} Gemini 3.1 Pro and Qwen3.8-27B cannot be separated on accuracy, yet their failure compositions are nearly opposite (\tabref{tab:profiles}, \figref{fig:profiles}):
\begin{itemize}
\item \textbf{Factual error.} Qwen3.8-27B records it on 46.4\% of captions [39.1, 53.7], ten points clear of any other model. Gemini 3.1 Pro is at 30.3\% [23.8, 37.4], mid-field.
\item \textbf{Segmentation.} Qwen3.8-27B has the lowest failure rate in the study, 6.8\% [3.7, 11.3]. Gemini 3.1 Pro has the highest, 18.1\% [12.9, 24.3].
\item \textbf{Omission.} Gemini 3.1 Pro is also highest, 23.9\%.
\item \textbf{Temporal order.} Gemini 3.1 Pro is highest at 2.1\%; Qwen3.8-27B records none.
\end{itemize}
Qwen3.8-27B asserts wrong content but places events well: it has the worst factual-error rate and the best timeline structure in the study. Gemini 3.1 Pro makes fewer factual errors but omits more and places events worse, and it writes the shortest captions (median overview 63.5 words, timeline 141), as if saying less to stay safe. \figref{fig:pair} sets the two profiles side by side. Camera is the flattest mode in relative terms: it spans 10.9--14.9\%, a max-to-min ratio of 1.4 against 1.7--2.7 for the other modes with non-zero floors. No model in the study is markedly better at it than the others. Factual error is the opposite, running from 24.9\% to 46.4\%.

\begin{table}[t]
\centering
\caption{Share of each model's captions carrying each mode at least once, detected window by window, over all 1,144 captions with a final verdict. Modes are not exclusive: 36\% of rejected captions carry two or more under this detector (16\% under the overview detector), so rows do not sum to a rejection rate. Appendix~\ref{app:paired} gives the paired-subset version.}
\label{tab:profiles}
\tablebox{\begin{tabular}{lccccccc}
\toprule
Model & $n$ & Factual & Omission & Camera & Inference & Temporal & Segment. \\
\midrule
Qwen3.8-Max & 193 & 24.9 & 16.6 & 10.9 & 4.7 & 0.5 & 12.4 \\
Muse Spark 1.2 & 187 & 36.4 & 14.4 & 13.9 & 4.3 & 0.5 & 14.4 \\
Kimi K3 & 194 & 33.5 & 13.9 & 11.3 & 5.2 & 1.0 & 10.8 \\
Seed 2.1 Turbo & 190 & 32.1 & 19.5 & 13.7 & 6.8 & 1.6 & 17.9 \\
Gemini 3.1 Pro & 188 & 30.3 & 23.9 & 14.9 & 4.8 & 2.1 & 18.1 \\
Qwen3.8-27B & 192 & 46.4 & 19.3 & 12.5 & 3.6 & 0.0 & 6.8 \\
\bottomrule
\end{tabular}
}
\end{table}

\paragraph{Scale within one family.} Qwen3.8-27B against Qwen3.8-Max is the only comparison in the study within one model series. On the paired subset the raw gap is 18.6 points. Adjusted for annotator and clip effects it is 20.7 points [11.1, 30.6], against an adjusted spread of 3.5 points across the three mid-table frontier models. Most of the difference is factual error (46.4\% against 24.9\%), while the smaller model records the study's best segmentation rate.

The comparison is not controlled for scale alone. The two models differ in more than size: one is an open dense model, the other a proprietary flagship whose architecture is not published. They were served by different stacks (ten third-party providers against one first-party provider). They received different input files on 11 clips. And Qwen3.8-27B was annotated in a separate, later batch, though by annotators who also reviewed the other models and who rejected the other models' captions at the corpus rate (31.3\% against 31.6\%). Within this series, size is associated with factual grounding and not with temporal structure, but the association is not a clean scaling result.

\subsection{Which component of the caption survives}

The mode taxonomy asks what kind of thing went wrong. A complementary cut asks which \emph{component} the annotator had to touch at all, which requires no taxonomy and no rejection threshold: either the component came back unchanged or it did not (\tabref{tab:components}).

\begin{table}[t]
\centering
\caption{Share of each component left untouched by the annotator, over all 1,144 captions. Denominators differ by column: overview and segmentation are per caption, scene and shot text per window. \emph{Rewritten}: share of all characters of overview and timeline text changed by the reviewer, with accepted text counted as unchanged (Appendix~\ref{app:definitions}).}
\label{tab:components}
\tablebox{\begin{tabular}{lccccc}
\toprule
Model & Overview & Scene text & Shot text & Segmentation & Rewritten \\
\midrule
Qwen3.8-Max & 72.5 & 83.7 & 96.1 & 87.6 & 2.78 \\
Muse Spark 1.2 & 71.7 & 74.0 & 91.6 & 85.6 & 2.94 \\
Kimi K3 & 71.1 & 77.9 & 94.4 & 89.2 & 3.67 \\
Seed 2.1 Turbo & 67.9 & 77.8 & 91.7 & 82.1 & 4.57 \\
Gemini 3.1 Pro & 58.5 & 70.2 & 88.5 & 81.9 & 10.62 \\
Qwen3.8-27B & 52.6 & 67.5 & 89.4 & 93.2 & 5.51 \\
\bottomrule
\end{tabular}
}
\end{table}

Columns should be compared down, not across, because the units differ in size. One overview covers a whole clip, so an error anywhere sinks it, while one shot line covers a few seconds and is often generic enough to survive by default. Shot text accordingly leads for every model (88.5--96.1\%) and overview trails (52.6--72.5\%), without that saying anything about which is written better. The rewritten share follows accuracy at the extremes, with Gemini 3.1 Pro highest (10.6\%) and Qwen3.8-Max lowest (2.8\%). In between it separates models with near-identical accuracy: Muse Spark 1.2's rejected text is concentrated in a minority of near-total rewrites, so its overall rewritten share (2.9\%) is low despite the highest critical rate.

The informative observation is the tension with \secref{sec:granularity}. Shot text is the \emph{least} frequently touched component, yet camera and shot is the mode most likely to need a near-total rewrite and the one the overview read barely detects. Shot description is therefore not a low-risk area. It is a low-frequency, high-cost one, and its low touch rate partly reflects how generic these lines are rather than how accurate. Qwen3.8-27B shows the same content-versus-structure split seen in its mode profile: the worst overview survival in the study, 52.6\%, alongside the best segmentation, 93.2\%.

\section{What the corrections indicate}
\label{sec:corrections}

Each window repair adds content the model omitted, removes a claim it should not have made, or substitutes one claim for another of similar length. \tabref{tab:direction} classifies each changed shot or scene field by whether the correction is longer, shorter, or within $\pm$5\% of the original length. The labels depend on the tolerance chosen, so the table also reports the share lengthened at $\pm$0\% and $\pm$10\%, and the median relative length change. The ordering is stable across tolerances. Gemini 3.1 Pro and Seed 2.1 Turbo have the most additive repairs at every tolerance (median length change $+$16.7\% and $+$8.2\%), Muse Spark 1.2 and Kimi K3 the least ($+$1.4\% and $+$2.4\%), and the two Qwen3.8 models sit between.

\begin{table}[t]
\centering
\caption{Direction of window repairs (changed shot or scene fields of rejected windows), over all 1,144 captions, ordered by the share lengthened at $\pm$5\%. \emph{Handheld penalty}: intervention rate on handheld minus tripod-static footage, from \tabref{tab:conditions}.}
\label{tab:direction}
\tablebox{\begin{tabular}{lcccccc}
\toprule
& & \multicolumn{3}{c}{Lengthen / shorten / same (\%)} & Median length & Handheld \\
\cmidrule(lr){3-5}
Model & Repairs & $\pm$5\% band & Lengthen, $\pm$0\% & Lengthen, $\pm$10\% & change (\%) & penalty (pts) \\
\midrule
Gemini 3.1 Pro & 310 & 65 / 14 / 22 & 72 & 56 & +16.7 & +3.1 \\
Seed 2.1 Turbo & 247 & 57 / 13 / 30 & 75 & 47 & +8.2 & +2.7 \\
Qwen3.8-27B & 330 & 50 / 19 / 31 & 64 & 42 & +5.6 & +26.1 \\
Qwen3.8-Max & 160 & 50 / 18 / 32 & 67 & 42 & +5.0 & +4.0 \\
Kimi K3 & 237 & 39 / 16 / 45 & 63 & 33 & +2.4 & +13.2 \\
Muse Spark 1.2 & 241 & 38 / 15 / 47 & 61 & 26 & +1.4 & +16.0 \\
\bottomrule
\end{tabular}
}
\end{table}

The distinction determines how a correction should be read. Where repairs are additive the model under-describes, and the corrected caption is a complete description of the clip in its own right. This fits Gemini 3.1 Pro, which writes the shortest overviews (median 63.5 words) and timelines (141 words) in the study. Where repairs are substitutive the model asserts wrong content in place, the correction alone is uninformative, and the diagnostic object is the pair: the incorrect claim beside the accurate one.

The handheld column suggests a second axis, explored in \secref{sec:sensitivity}. Qwen3.8-27B, Muse Spark 1.2 and Kimi K3 lose 26.1, 16.0 and 13.2 points on handheld relative to tripod-static footage. The other three are approximately flat, and the split does not follow accuracy.

\tabref{tab:model_profiles} brings each model's accuracy, most frequent modes, handheld penalty and repair direction together. We label a model \emph{additive} when at least 55\% of its repairs lengthen the text beyond the $\pm$5\% band, \emph{substitutive} when fewer than 45\% do, and \emph{mixed} otherwise. These labels move with the tolerance (\tabref{tab:direction}); the ordering behind them does not.

\begin{table}[!htbp]
\centering
\caption{Model profiles. Accuracy: correct-overview rate on the 172-clip paired subset. Modes: the two most frequent, window detector, all captions (\tabref{tab:profiles}). Handheld penalty: points lost on handheld against tripod-static footage (\tabref{tab:conditions}). Repairs: share of window repairs that lengthen, shorten or keep the length within $\pm$5\% (\tabref{tab:direction}). Direction: additive $\geq$55\% lengthened, substitutive $<$45\%, mixed otherwise.}
\label{tab:model_profiles}
\tablebox{\begin{tabular}{@{}lc>{\raggedright\arraybackslash}p{5.4cm}ccc@{}}
\toprule
Model & Accuracy (\%) & Two most frequent modes (\%) & Handheld & Repairs L / S / = & Direction \\
\midrule
Qwen3.8-Max & 72.7 & factual 24.9; omission 16.6 & +4.0 & 50 / 18 / 32 & mixed \\
Muse Spark 1.2 & 72.1 & factual 36.4; segmentation = omission 14.4 & +16.0 & 38 / 15 / 47 & substitutive \\
Kimi K3 & 71.5 & factual 33.5; omission 13.9 & +13.2 & 39 / 16 / 45 & substitutive \\
Seed 2.1 Turbo & 69.8 & factual 32.1; omission 19.5 & +2.7 & 57 / 13 / 30 & additive \\
Gemini 3.1 Pro & 58.7 & factual 30.3; omission 23.9 & +3.1 & 65 / 14 / 22 & additive \\
Qwen3.8-27B & 54.1 & factual 46.4; omission 19.3 & +26.1 & 50 / 19 / 31 & mixed \\
\bottomrule
\end{tabular}
}
\end{table}

\paragraph{What the profiles suggest for training data.} The profiles point to different kinds of training data. We have not tested any of these interventions (\secref{sec:limitations}), so they are hypotheses for follow-up rather than findings.
\begin{itemize}
\item \textbf{Contrastive correction pairs.} Where repairs mostly replace wrong claims in place (Muse Spark 1.2 and Kimi K3), the informative signal is the incorrect claim set beside what the footage shows. The same applies to Qwen3.8-27B: its repair direction is mixed, but its profile is dominated by factual error (46.4\%).
\item \textbf{Complete, correctly timed captions.} Where repairs mostly add (Gemini 3.1 Pro and Seed 2.1 Turbo), the informative signal is the corrected caption itself, which shows what ``complete'' and ``correctly placed'' look like. Gemini 3.1 Pro also has the highest omission, segmentation and temporal-order rates in the study.
\end{itemize}
Separately, a conditional weakness identifies footage to collect more of (handheld clips for Qwen3.8-27B), while a flat one does not.

\section{Which footage is difficult}
\label{sec:difficulty}

Pooling all six models over the 1,144 captions removes the model from the comparison and isolates properties of the clip. Because the properties are measured per clip and each clip contributes up to six captions, we test them with clips as the unit: a Kruskal--Wallis test on each clip's correct-overview share. \tabref{tab:properties} also reports the caption-level $\chi^2$ for comparison. With six properties tested, the Bonferroni threshold is $p<0.0083$.

\begin{table}[t]
\centering
\caption{Correct-overview rate pooled across the six models, by clip property (labels inferred by a language model; \secref{sec:protocol}). Buckets with fewer than 60 verdicts and buckets labelled \emph{unclear} are omitted. Caption-level $\chi^2$ treats captions as independent and overstates the evidence. The clip-level test (Kruskal--Wallis on per-clip correct shares) is the appropriate one; $^\ast$ survives Bonferroni correction across the six properties.}
\label{tab:properties}
\tablebox{\begin{tabular}{llcccc}
\toprule
& & \multicolumn{3}{c}{Caption level} & Clip level \\
\cmidrule(lr){3-5}\cmidrule(lr){6-6}
Property & Correct-overview range across buckets & $\chi^2$ & d.f. & $p$ & $p$ \\
\midrule
Content type & 54.5 domestic lifestyle -- 72.6 people activity & 21.6 & 4 & $<$0.001 & 0.001$^\ast$ \\
Camera platform & 56.2 vehicle mounted -- 72.1 tripod static & 15.2 & 5 & 0.010 & 0.015 \\
Scene complexity & 61.3 medium -- 69.0 low & 7.1 & 1 & 0.008 & 0.011 \\
Shot scale & 60.4 mixed -- 80.0 extreme wide & 10.6 & 4 & 0.032 & 0.039 \\
Motion intensity & 62.1 medium -- 69.9 high & 3.4 & 2 & 0.180 & 0.207 \\
Setting & 63.5 indoor -- 66.5 outdoor & 1.1 & 1 & 0.300 & 0.293 \\
\bottomrule
\end{tabular}
}
\end{table}

Only content type survives at clip level ($p=0.001$). Domestic and lifestyle footage scores 18.1 points below people-and-activity footage on comparably sized samples. Camera platform ($p=0.015$) and scene complexity ($p=0.011$) are suggestive but do not survive correction; a locked-down rig yields 72.1\% while handheld, POV and vehicle-mounted footage fall between 56\% and 62\%. The complexity effect is not monotone. Medium-complexity clips score lowest, while the small high-complexity bucket (48 verdicts, 68.8\%) scores as well as low-complexity clips, so ``clutter'' is not an adequate description. Shot scale, motion intensity and setting show no reliable effect. Indoor and outdoor footage differ by 3.0 points, and shot scale ranges from 80.0\% on extreme-wide shots to 60.4\% on clips mixing scales. We expected fast footage to be harder; it is not, and high-motion clips score above medium-motion ones (69.9\% against 62.1\%). Motion alone is therefore not a reliable difficulty signal, though we would not conclude that fast footage is easier.

How hard a slice is and how well it separates models are different properties. \tabref{tab:buckets} breaks content type and camera platform into buckets and adds, for each, the spread between the best and worst model. Vehicle-mounted footage scores 15.8 points below tripod-static. The vehicle-mounted, body-worn, gimbal-stabilised and drone buckets show the widest spreads: gimbal-stabilised footage puts 63.6 points between the best and worst model, against 23.5 for people-and-activity footage. Those buckets are small, however: 11 to 14 clips, or about 11 captions per model, so each caption moves a model's rate by about 9 points. Their spreads are indicative only. Handheld footage, the largest moving-camera bucket at 64 clips, has one of the narrowest spreads (24.3). The small moving-camera slices are the thinnest in this corpus and, on this evidence, among the most discriminating.

\begin{table}[!htbp]
\centering
\caption{Correct-overview rate by bucket for the two properties with the largest effects, pooled across the six models, all captions. \emph{Per model}: range of captions per model in the bucket. \emph{Model spread}: best minus worst model's correct-overview rate in the bucket. Buckets with fewer than 60 verdicts and \emph{unclear} labels are omitted.}
\label{tab:buckets}
\tablebox{\begin{tabular}{lccccc}
\toprule
Bucket & Correct (\%) & Verdicts & Clips & Per model & Model spread (pts) \\
\midrule
\multicolumn{6}{l}{\emph{Content type}} \\
\quad domestic lifestyle & 54.5 & 233 & 40 & 36--40 & 31.7 \\
\quad urban vehicle & 63.6 & 231 & 40 & 38--39 & 25.0 \\
\quad nature outdoor & 66.9 & 236 & 40 & 38--40 & 30.0 \\
\quad product object & 71.5 & 207 & 36 & 32--36 & 23.8 \\
\quad people activity & 72.6 & 237 & 40 & 39--40 & 23.5 \\
\addlinespace
\multicolumn{6}{l}{\emph{Camera platform}} \\
\quad vehicle mounted & 56.2 & 64 & 11 & 10--11 & 45.5 \\
\quad pov body worn & 57.7 & 71 & 12 & 11--12 & 50.0 \\
\quad handheld & 61.1 & 373 & 64 & 60--64 & 24.3 \\
\quad gimbal stabilized & 63.1 & 65 & 11 & 10--11 & 63.6 \\
\quad drone aerial & 70.7 & 82 & 14 & 13--14 & 35.7 \\
\quad tripod static & 72.1 & 344 & 59 & 55--59 & 29.6 \\
\bottomrule
\end{tabular}
}
\end{table}

\begin{figure}[t]
\centering
\includegraphics[width=0.92\linewidth]{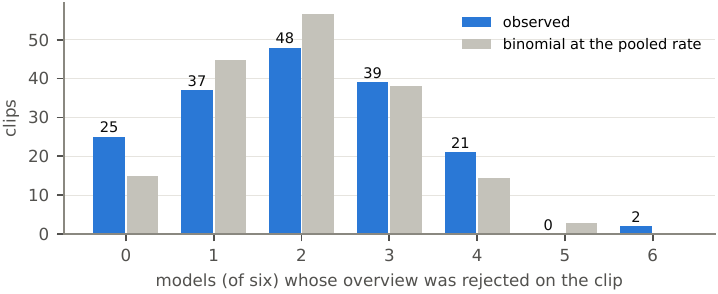}
\caption{Number of the six models failing each of the 172 paired clips, against a binomial at the pooled rate. The spread is wider than independence predicts, but the tails are thin: two clips defeated every model and 25 defeated none.}
\label{fig:cofailure}
\end{figure}

Difficulty is mostly model-specific rather than intrinsic to the footage (\figref{fig:cofailure}). On the paired subset, the number of models failing each clip is more dispersed than independence predicts: a Monte Carlo test of the variance against independent failures at each model's own rate gives $p=0.002$. But the shared component is small, with an intraclass correlation of 0.055. Two clips defeated all six models and 25 of 172 defeated none.

\subsection{Sensitivity to difficult footage differs by model}
\label{sec:sensitivity}

\begin{figure}[t]
\centering
\includegraphics[width=\linewidth]{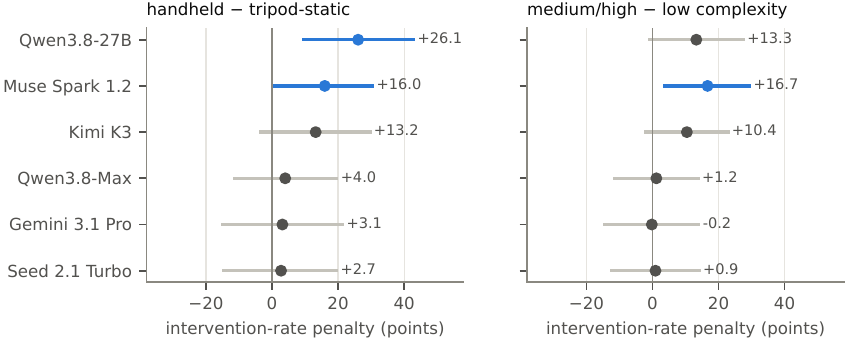}
\caption{Penalty on difficult footage with 95\% bootstrap intervals over clips. Blue marks intervals that exclude zero. Models are sorted by handheld penalty.}
\label{fig:conditions}
\end{figure}

Whether a \emph{given} model is unusually hurt by a condition is a separate question from what is hard for the corpus. \tabref{tab:conditions} and \figref{fig:conditions} give each model's penalty on handheld against tripod-static footage and on medium/high against low scene complexity, with bootstrap intervals over clips. Each arm holds about 60 captions per model, so this analysis is exploratory.
\begin{itemize}
\item \textbf{Clear penalties.} Qwen3.8-27B's handheld penalty (+26.1 points [9.5, 42.9]) and Muse Spark 1.2's penalties on both conditions have intervals excluding zero.
\item \textbf{Inconclusive.} Kimi K3's handheld penalty (+13.2) does not exclude zero.
\item \textbf{Approximately flat.} Qwen3.8-Max, Gemini 3.1 Pro and Seed 2.1 Turbo, with point estimates within 4 points of zero on both conditions.
\item \textbf{Model differences.} The largest difference between two models' handheld penalties, Qwen3.8-27B against Qwen3.8-Max, has interval [0.3, 44.0], and it is uncorrected for the other comparisons that could have been made.
\end{itemize}

\begin{table}[t]
\centering
\caption{Intervention rate (the complement of the correct-overview rate) on difficult against easier footage, per model, over all 1,144 captions. $\Delta$ is the penalty in points, with a 95\% bootstrap interval over clips. Complexity compares medium or high against low.}
\label{tab:conditions}
\tablebox{\begin{tabular}{lcccccc}
\toprule
& \multicolumn{3}{c}{Camera platform} & \multicolumn{3}{c}{Scene complexity} \\
\cmidrule(lr){2-4}\cmidrule(lr){5-7}
Model & Handheld & Tripod & $\Delta$ [95\% CI] & Med./high & Low & $\Delta$ [95\% CI] \\
\midrule
Qwen3.8-27B & 52.4 & 26.3 & +26.1 [9.5, 42.9] & 54.4 & 41.2 & +13.3 [$-$0.8, 27.6] \\
Muse Spark 1.2 & 30.0 & 14.0 & +16.0 [0.9, 30.5] & 37.1 & 20.4 & +16.7 [3.8, 29.4] \\
Kimi K3 & 40.3 & 27.1 & +13.2 [$-$3.5, 29.6] & 34.4 & 24.0 & +10.4 [$-$2.2, 22.9] \\
Qwen3.8-Max & 28.1 & 24.1 & +4.0 [$-$11.2, 19.6] & 28.1 & 26.9 & +1.2 [$-$11.6, 13.8] \\
Gemini 3.1 Pro & 46.8 & 43.6 & +3.1 [$-$14.9, 21.4] & 41.4 & 41.6 & $-$0.2 [$-$14.4, 14.0] \\
Seed 2.1 Turbo & 35.5 & 32.8 & +2.7 [$-$14.7, 19.5] & 32.6 & 31.7 & +0.9 [$-$12.4, 14.1] \\
\bottomrule
\end{tabular}
}
\end{table}

With that caveat, two qualitatively different patterns appear. A \emph{conditional} weakness, as for Qwen3.8-27B and Muse Spark 1.2, suggests a specific condition to sample more heavily when commissioning annotation. A \emph{flat} profile is a different case, and it splits in two:
\begin{itemize}
\item \textbf{Flat and high:} Gemini 3.1 Pro, a uniform weakness that targeted collection will not fix.
\item \textbf{Flat and low:} Qwen3.8-Max, which is what robustness looks like.
\end{itemize}
A single pooled difficulty figure would report the same thing for Gemini 3.1 Pro and Qwen3.8-Max despite these being opposite situations. Motion intensity is noisier still. Seed 2.1 Turbo and Qwen3.8-Max score better on high-motion than on low-motion footage ($-17.3$ and $-10.0$ points), on about 20 high-motion clips each, so we do not interpret it.

\section{Hypotheses that did not survive}
\label{sec:negative}

We report these because a reader planning similar work would otherwise spend budget testing them again.

\begin{figure}[t]
\centering
\includegraphics[width=0.6\linewidth]{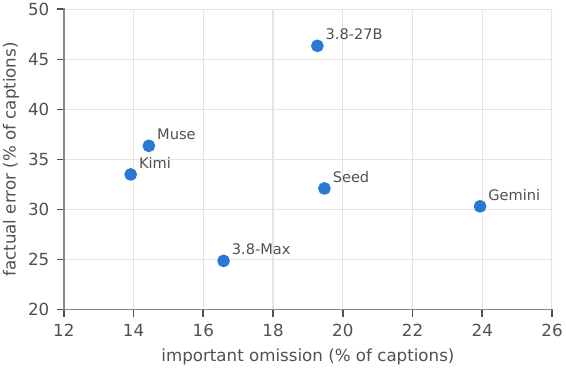}
\caption{Omission rate against factual-error rate, one point per model, window detector, all captions. A precision--recall trade-off would place these on a descending line; the correlation is $r=-0.02$.}
\label{fig:tradeoff}
\end{figure}

\paragraph{No precision--recall trade-off, and a trap worth naming.} The intuition that a model omitting more should invent less does not hold (\figref{fig:tradeoff}). Across the six models, omission rate and factual-error rate correlate at $r=-0.02$ with the window detector and $r=+0.71$ with the overview detector. Neither is the negative relation a trade-off predicts. Within models, factual error is more common in captions that also omit for four of six models, markedly so for Qwen3.8-27B (67.6\% against 41.3\%) and Muse Spark 1.2 (51.9\% against 33.8\%), and indistinguishable for Kimi K3 (33.3\% against 33.5\%) and Gemini 3.1 Pro (31.1\% against 30.1\%).

The trap is worth stating because it produces a clean, publishable and wrong result. Computing mode co-occurrence on rejected windows only yields $\phi=-0.52$ (odds ratio 0.07), an apparently strong trade-off. It is an artefact of conditioning on a collider \citep{berkson1946,elwert2014collider}. Modes are recorded only where a unit was rejected, and a unit is rejected \emph{because} it carries at least one mode, so within the rejected set the presence of one mode makes another less necessary to explain the rejection. Over all 4,672 judged windows the same association is $\phi=-0.01$ (odds ratio 0.88). Over captions it is $\phi=+0.07$ (odds ratio 1.45; 1.36 stratified by clip). Even the full-denominator figures remain censored, since accepted units carry no labels by construction. The supportable conclusion is the absence of a trade-off, not a positive association. Any co-occurrence or trade-off question on data of this shape must use the full denominator; restricted to rejected units it will manufacture a negative association whatever the underlying truth.

\paragraph{Caption-level and window-level rankings agree.} An interim analysis of 73 paired clips suggested they did not ($\rho=-0.03$). On the full 172 clips $\rho=+0.83$ (exact $p=0.058$ with six models). The inversion was a small-sample artefact. What survives is the weaker claim that the \emph{size} of each model's gap between the two levels differs (\tabref{tab:primary}).

\paragraph{Neither drift nor window length explains window error.} Error rate by quartile of position within the clip runs 31.6, 29.8, 27.9 and 27.1\%, drifting mildly in the direction opposite to accumulating error. Rejected and accepted windows have the same median duration, 3.5\,s.

\paragraph{Longer captions do not fail more.} A model might fail more simply because it asserts more. Across models, median overview length and intervention rate are unrelated (Spearman $\rho=+0.09$). Within each model, the correlation between overview length and rejection lies between $0.00$ and $+0.14$.

\paragraph{A video-LLM judge cannot yet audit reviewers.} We ran Gemini 3.1 Pro, with video, to list the material defects in each original caption, then to compare its list with the annotator's edits; Kimi K3 repeated this on 130 captions as a cross-check. The judge's defect count separates accepted from rejected captions only weakly (AUC 0.558, \tabref{tab:metrics}). The two judges agree poorly on whether a reviewer missed a defect ($\kappa=0.25$). Each finds the fewest defects in its own family's captions: Gemini 3.1 Pro finds a mean of 1.19 defects per Gemini caption against 1.64--2.48 for the other models, and Kimi K3 finds 0.48 per Kimi caption against 0.74. We therefore use no judge output in any verdict reported here, and treat judge-flagged ``missed defects'' as unverified.

\section{Limitations}
\label{sec:limitations}

\paragraph{One annotator per caption.} Each caption was post-edited by one annotator and checked by a QC reviewer who saw that work. The QC pass changed few verdicts, but it is not an independent second annotation, so we cannot report inter-annotator agreement. Agreement on accept/reject, and on the set of modes (for which a set-valued reliability measure such as Krippendorff's $\alpha$ \citep{hayes2007krippendorff} would be appropriate), remains to be measured on an independently double-annotated subset.

\paragraph{Severity is derived.} The annotation platform did not capture a severity field, so ``critical'' and ``near-total rewrite'' are computed from the size of the correction rather than assigned by the reviewer. Edit size and error severity are different things \citep{vanmiltenburg2020gradations}: a one-word substitution that reverses a claim scores as minor. Every critical rate here is therefore a lower bound, and the ordering across models and modes is more reliable than the absolute levels.

\paragraph{Mode incidence is a floor.} Failure modes were recorded only on rejected units, so a mode present within an accepted unit was never logged.

\paragraph{Reviewer assignment was not blocked.} Annotators did not all see the same mix of clips and models, and two of them reviewed a third of the data. After removing binomial sampling noise, the standard deviation of the rejection rate is 11.1 points across clips, 7.3 across annotators and 7.1 across models. Raw spreads overstate the first two, because a clip carries only six verdicts. The two large model gaps survive the adjustment: against Qwen3.8-Max, Qwen3.8-27B is +20.7 [11.1, 30.6] and Gemini 3.1 Pro +13.2 [3.7, 23.3]. The latter does not survive Bonferroni correction, and differences of a few points in the middle of the table should not be read as an ordering.

\paragraph{Qwen3.8-27B was annotated later.} It was added after the other models and reviewed on the last two days of annotation. The same annotators reviewed it, and their strictness on the other models matches the corpus rate, but a drift in standards over time cannot be excluded.

\paragraph{Inputs differed across providers.} Frame sampling and resolution were left to each provider, the Qwen3.8 models received re-encoded copies of oversized clips, and invalid outputs were retried with validator feedback (\secref{sec:protocol}). The comparison is of systems as served, not of models on identical inputs.

\paragraph{Metadata and audio.} The clip properties of \secref{sec:difficulty} were inferred by a language model from text, not labelled by humans from the video. Models received muted video, but annotators were shown the original files: 57 of 196 carry an audio track and 15 are audibly non-silent. The rubric did not instruct annotators to ignore sound, so a small amount of audio information may have entered their judgements.

\paragraph{Corpus composition.} The corpus mixes stock, first-person household and internally sourced footage, 54 clips have no recorded origin, and 12 clips are manipulated versions of source footage. Excluding the manipulated clips leaves the paired accuracy range at 53.7--72.2\%, the number of separable model pairs at 4, and the camera ratio at 4.5$\times$.

\paragraph{Scope.} The corpus is 196 short clips, overwhelmingly under thirty seconds, evaluated with a single prompt. Nothing here extrapolates to long-form video, to a different domain or to a different prompt.

\paragraph{One system was excluded, not tested and passed over.} Qwen3.5-397B was captioned and reviewed on the same protocol, but its outputs pool two generation runs differing by about seven points, so no single row describes it. The seven-model corpus comprises 1,372 planned captions, 1,342 annotated, 1,331 with a final verdict (454 of them rejected with corrections) and 5,431 timeline windows. Including the seventh model gives a 165-clip paired subset, a correct-overview range of 52.7--73.9\%, and a camera ratio of 4.6$\times$. Six of 21 raw pairwise gaps survive correction there. Two of those six do not carry over to the six-model analysis. Qwen3.5-397B against Qwen3.8-27B leaves with the model. Gemini 3.1 Pro against Qwen3.8-Max (raw gap 14.0 on the 172 clips, Bonferroni interval $-1.2$--28.5) no longer survives. This leaves 4 of 15, and Gemini 3.1 Pro separable from Muse Spark 1.2 only after adjustment.

\paragraph{Not measured.} No timing or cost data was collected, so no claim about review effort or saving is supported. This study does not establish that correcting a model's outputs improves that model; the intervention experiment that would test it has not been run.

\section{Conclusion}

Caption evaluation currently returns a scalar. On dense video description that scalar is difficult to obtain reliably: overlap metrics do not recover human judgement on this corpus, and even when obtained from human review it conceals the structure that determines what to do next. Two models within each other's confidence intervals differed sharply in composition. A model second on accuracy was last on severity. And the mode most likely to need a near-total rewrite was the one an overview-level read was least likely to record.

The alternative we report is not a better scalar. It is to have annotators post-edit rather than grade, to type every failure against a taxonomy fixed in advance, and to judge output at more than one granularity. What that yields is a record of 1,144 overview verdicts, corrections for the 392 rejected overviews and 4,672 window-level judgements, from which a mode can be scored as its own slice rather than absorbed into an average. Because every error is typed, later evaluations can ask narrower questions than ``which model wins'': whether a model improved at camera understanding, whether it misses fewer important events, whether its windows sit where the events are.

The method is not specific to these models or this footage. The numbers are.

\paragraph{Reproducibility and data availability.}\begin{sloppypar}Every number, table and figure in this paper is regenerated from the annotation export by the accompanying analysis scripts (\texttt{table1.py}, \texttt{paper\_numbers.py}, \texttt{tables.py}, \texttt{figures.py}); the definitions of all derived measures are given in Appendix~\ref{app:definitions}.
Model captions, reviewer corrections, mode labels and clip identifiers are available on request at \url{https://data.deccan.ai}, and the analysis scripts accompany this paper as ancillary files. The 107 Pexels clips are identified by their public source. The remaining 89 clips cannot be redistributed.\end{sloppypar}

\bibliographystyle{plainnat}
\bibliography{refs}

\appendix

\section{Model access and generation runs}
\label{app:runs}

\begin{table}[H]
\centering
\caption{Generation details. \emph{Providers}: distinct serving providers recorded in the responses. \emph{Tokens/s video}: median input tokens per second of video on first attempts, a proxy for each provider's frame sampling. \emph{Retried}: clips re-requested with the schema validator's message appended. \emph{Re-encoded}: clips sent as a different file from the one the other models received (the Qwen3.8 endpoints' 6\,MB limit). \emph{Invalid kept}: outputs reviewed despite never passing validation. All runs used temperature 0 and muted video, between 16 and 27 August 2026.}
\label{tab:runs}
\tablebox{\begin{tabular}{llcccccc}
\toprule
Model & Identifier & Route & Providers & Tokens/s video & Retried & Re-encoded & Invalid kept \\
\midrule
Qwen3.8-Max & \texttt{qwen/qwen3.8-max} & OpenRouter & 1 & 739 & 38 & 24 & 5 \\
Muse Spark 1.2 & \texttt{meta/muse-spark-1.2} & OpenRouter & 1 & 283 & 0 & 0 & 0 \\
Kimi K3 & \texttt{moonshotai/kimi-k3} & OpenRouter & 1 & 2596 & 55 & 0 & 0 \\
Seed 2.1 Turbo & \texttt{bytedance-seed/seed-2-1-turbo} & OpenRouter & 1 & 754 & 7 & 0 & 0 \\
Gemini 3.1 Pro & \texttt{gemini-3.1-pro-preview} & Gemini API & 1 & 211 & 0 & 0 & 0 \\
Qwen3.8-27B & \texttt{qwen/qwen3.8-27b} & OpenRouter & 10 & 775 & 22 & 35 & 0 \\
\bottomrule
\end{tabular}
}
\end{table}

\section{Pairwise model comparisons}
\label{app:pairwise}

\begin{table}[H]
\centering
\caption{Pairwise gaps in overview intervention rate (row model minus column model, points) on the 172-clip paired subset. Raw: paired bootstrap over clips, 20,000 resamples. Adjusted: additive model with annotator and clip effects refitted in each of 4,000 clip-bootstrap resamples. Intervals are Bonferroni-corrected for 15 comparisons.}
\label{tab:pairwise}
\small
\tablebox{\begin{tabular}{lcccccc}
\toprule
& \multicolumn{3}{c}{Raw} & \multicolumn{3}{c}{Reviewer- and clip-adjusted} \\
\cmidrule(lr){2-4}\cmidrule(lr){5-7}
Pair (row $-$ column) & Gap & Bonferroni CI & Separates & Gap & Bonferroni CI & Separates \\
\midrule
3.8-Max -- Muse & $-$0.6 & [$-$15.1, 13.4] &  & +0.9 & [$-$13.3, 15.2] &  \\
3.8-Max -- Kimi & $-$1.2 & [$-$14.5, 12.2] &  & $-$2.6 & [$-$15.1, 10.3] &  \\
3.8-Max -- Seed & $-$2.9 & [$-$16.3, 11.0] &  & $-$1.2 & [$-$15.0, 12.8] &  \\
3.8-Max -- Gemini & $-$14.0 & [$-$28.5, 1.2] &  & $-$13.2 & [$-$27.8, 2.3] &  \\
3.8-Max -- 3.8-27B & $-$18.6 & [$-$32.6, $-$4.7] & yes & $-$20.7 & [$-$35.6, $-$6.6] & yes \\
Muse -- Kimi & $-$0.6 & [$-$14.5, 13.4] &  & $-$3.5 & [$-$18.1, 10.3] &  \\
Muse -- Seed & $-$2.3 & [$-$16.3, 11.6] &  & $-$2.1 & [$-$16.6, 12.1] &  \\
Muse -- Gemini & $-$13.4 & [$-$27.3, 1.0] &  & $-$14.1 & [$-$27.8, $-$0.8] & yes \\
Muse -- 3.8-27B & $-$18.0 & [$-$31.4, $-$4.7] & yes & $-$21.6 & [$-$34.3, $-$9.0] & yes \\
Kimi -- Seed & $-$1.7 & [$-$16.1, 12.8] &  & +1.4 & [$-$12.9, 16.4] &  \\
Kimi -- Gemini & $-$12.8 & [$-$25.6, 0.6] &  & $-$10.6 & [$-$24.0, 1.9] &  \\
Kimi -- 3.8-27B & $-$17.4 & [$-$32.0, $-$2.9] & yes & $-$18.1 & [$-$32.5, $-$4.3] & yes \\
Seed -- Gemini & $-$11.0 & [$-$26.7, 4.7] &  & $-$12.0 & [$-$28.4, 3.4] &  \\
Seed -- 3.8-27B & $-$15.7 & [$-$30.2, $-$1.2] & yes & $-$19.5 & [$-$34.1, $-$5.7] & yes \\
Gemini -- 3.8-27B & $-$4.7 & [$-$20.2, 11.0] &  & $-$7.5 & [$-$22.7, 7.2] &  \\
\bottomrule
\end{tabular}
}
\end{table}

\section{Paired-subset versions of the descriptive tables}
\label{app:paired}

\begin{table}[H]
\centering
\caption{\tabref{tab:threshold} on the 172-clip paired subset.}
\tablebox{\begin{tabular}{lccccc}
\toprule
Model & $n$ & Any correction & $\geq$3 words & $\geq$6 words & $\geq$16 words \\
\midrule
Qwen3.8-Max & 172 & 27.3 & 22.1 & 16.9 & 7.0 \\
Muse Spark 1.2 & 172 & 27.9 & 23.8 & 19.2 & 7.6 \\
Kimi K3 & 172 & 28.5 & 23.3 & 19.8 & 7.6 \\
Seed 2.1 Turbo & 172 & 30.2 & 24.4 & 20.3 & 11.0 \\
Gemini 3.1 Pro & 172 & 41.3 & 35.5 & 27.9 & 15.1 \\
Qwen3.8-27B & 172 & 45.9 & 37.8 & 26.7 & 11.0 \\
\bottomrule
\end{tabular}
}
\end{table}

\begin{table}[H]
\centering
\caption{\tabref{tab:profiles} on the 172-clip paired subset.}
\tablebox{\begin{tabular}{lccccccc}
\toprule
Model & $n$ & Factual & Omission & Camera & Inference & Temporal & Segment. \\
\midrule
Qwen3.8-Max & 172 & 25.0 & 16.9 & 9.9 & 4.7 & 0.6 & 13.4 \\
Muse Spark 1.2 & 172 & 36.6 & 15.1 & 13.4 & 4.7 & 0.6 & 15.1 \\
Kimi K3 & 172 & 33.7 & 12.2 & 12.2 & 4.7 & 1.2 & 10.5 \\
Seed 2.1 Turbo & 172 & 32.6 & 19.2 & 13.4 & 5.8 & 1.7 & 18.0 \\
Gemini 3.1 Pro & 172 & 30.8 & 23.8 & 14.5 & 3.5 & 1.7 & 16.9 \\
Qwen3.8-27B & 172 & 44.2 & 19.8 & 12.8 & 4.1 & 0.0 & 7.6 \\
\bottomrule
\end{tabular}
}
\end{table}

\section{Definitions of derived measures}
\label{app:definitions}

\begin{itemize}
\item \textbf{Character-level similarity.} \texttt{difflib.SequenceMatcher(None, a, b).ratio()} in Python, i.e.\ $2M/T$, where $M$ is the number of matched characters and $T$ the total length of both strings.
\item \textbf{Critical caption (\tabref{tab:primary}).} At least one rejected window whose corrected text (shot and scene concatenated, uncorrected fields kept) has similarity below 0.55 to the original. \emph{Moderate}: similarity between 0.55 and 0.85.
\item \textbf{Near-total rewrite (\tabref{tab:cost}).} The same threshold, applied to rejected windows carrying the given mode.
\item \textbf{Changed words (\tabref{tab:threshold}).} Word-level diff of the overview against its correction: inserted plus deleted words, with replacements counted at the longer side.
\item \textbf{Rewritten (\tabref{tab:components}).} Changed characters (same diff rule, character level) summed over overview and every shot and scene field, divided by the total characters of the original text, with accepted text contributing zero changed characters.
\item \textbf{Repair direction (\tabref{tab:direction}).} Each changed shot or scene field of a rejected window is classed as lengthened or shortened if its corrected length differs from the original by more than the tolerance, and otherwise as same length.
\item \textbf{Hidden failure.} An accepted overview whose caption has at least one rejected window.
\end{itemize}

\end{document}